\documentclass[11pt]{article}

\usepackage[T1]{fontenc}
\usepackage[utf8]{inputenc}
\usepackage{palatino}
\usepackage[margin=1in]{geometry}
\usepackage{microtype}
\usepackage{graphicx}
\usepackage{booktabs}
\usepackage{array}
\usepackage{amsmath}
\usepackage{enumitem}
\usepackage{caption}
\usepackage{fancyhdr}
\usepackage[numbers,sort&compress]{natbib}
\usepackage[hidelinks]{hyperref}
\usepackage{xurl}
\usepackage{titlesec}
\usepackage{placeins}

\setlist[itemize]{leftmargin=1.35em,itemsep=0.1em,topsep=0.2em}

\titleformat{\section}
  {\large\bfseries}
  {\thesection}
  {0.6em}
  {}
\titleformat{\subsection}
  {\bfseries}
  {\thesubsection}
  {0.55em}
  {}
\titlespacing*{\section}{0pt}{1.2em}{0.35em}
\titlespacing*{\subsection}{0pt}{0.9em}{0.15em}

\graphicspath{{./}}

\hypersetup{
  pdftitle={Hard Cases, Bad Labels},
  pdfauthor={John Myron Uy},
  pdfsubject={Testing error exposure and error location in uncertainty sampling under bounded label noise}
}

\title{\textbf{Hard Cases, Bad Labels}\\
\large Testing Error Exposure and Error Location in Uncertainty Sampling
Under Bounded Label Noise}
\author{
  John Myron Uy\\
  \small Independent Researcher\\
  \small \texttt{johnmyron.uy@gmail.com}
}
\date{July 2026}

\begin{document}

\maketitle

\begin{abstract}
Active learning can reduce labeling cost by requesting examples that appear
informative, but the most uncertain examples may also be the hardest to label
correctly. This motivates two distinct explanations for failure: uncertainty
sampling may acquire more corrupted labels, and errors concentrated in
difficult regions may remain more harmful even after acquired corruption is
aligned. This study tests both explanations by comparing margin-based
uncertainty sampling with random sampling under clean labels, random
classification noise (RCN), and bounded difficulty-dependent noise on three
public binary tabular datasets. The design uses 100 paired seeds, nine
expected noise rates from 0 to 0.30, annotation budgets from 20 to 120, and
logistic regression with regularization re-selected by cross-validation at
every budget. A leave-one-seed-out exposure-matched RCN control tests whether
performance differences remain after aligning mean final acquired corruption;
a clean-label extension reaches budget 400. Under clean labels, uncertainty
sampling improved normalized balanced-accuracy area under the learning curve
(AULC) by 1.09 to 1.77 percentage points on all three datasets, with every
comparison surviving Holm correction. Difficulty-dependent noise reduced this
advantage more than RCN at six of eight rates on Breast Cancer Wisconsin, but
at no tested rate on Banknote Authentication or MAGIC Gamma Telescope. The
exposure-matched analysis found no Holm-corrected evidence for the predicted
additional harm of structured error location; two MAGIC comparisons were
significant in the opposite direction. No mean AULC advantage crossed below
zero through 0.30 noise. On clean MAGIC data, uncertainty sampling improved
balanced accuracy while reducing average precision and true-positive rate at
fixed false-positive rates. Uncertainty sampling was label-efficient, but its
apparent robustness depended on dataset, budget, noise structure, and metric.
\end{abstract}

\section{Introduction}

Many machine-learning projects begin with more unlabeled examples than can be
reviewed by experts. Active learning addresses this problem by allowing a
model to choose which examples should be labeled next \citep{settles2009}. A
common method, margin-based uncertainty sampling, requests examples closest
to the learner's current decision boundary \citep{lewis1994}. These examples
can be informative because their labels may change where the boundary is
placed.

The same examples can also be difficult for annotators. Human label quality
may vary by annotator, domain, and individual example
\citep{yan2011,cheng2021,frenay2014}.
If uncertainty sampling repeatedly requests ambiguous cases, it may acquire a
higher concentration of incorrect labels than random sampling. This creates a
tension: the examples with the greatest potential information may also carry
the greatest annotation risk.

This tension contains two mechanisms. First, \emph{exposure}: uncertainty
sampling may acquire a larger fraction of corrupted labels because it queries
where error probability is elevated. Second, \emph{location}: after acquired
corruption is aligned, a structured placement of errors may still be more
damaging than a random placement. Standard end-to-end comparisons combine
these mechanisms. Distinguishing them matters because exposure can motivate
repeated labeling or escalation, whereas a residual location effect would
implicate the acquisition rule itself.

This study provides a controlled paired benchmark on three binary tabular
datasets. It distinguishes constant-probability random classification noise
(RCN) from a bounded difficulty-dependent process whose error probability
increases near a fixed simulation-only reference boundary. It then compares
the structured process with an independent RCN condition calibrated to align
mean final acquired corruption. This control does not perfectly isolate
location within each seed or across the full trajectory; it tests whether a
residual performance difference remains after exposure alignment. The goal is
not to reproduce all human disagreement or propose a new query strategy. It is
to measure when a simple uncertainty baseline helps, when its advantage
changes, and which explanations survive replication across datasets.

\subsection{Contributions}

This study makes four bounded contributions:

\begin{itemize}
  \item It evaluates uncertainty and random sampling across complete
  low-budget learning curves using paired seeds and budget-specific
  regularization selection.
  \item It compares constant-probability RCN with a reproducible,
  difficulty-dependent noise process while reporting results separately by
  dataset.
  \item It uses a leave-one-seed-out exposure-matched control to test whether
  performance differences remain after aligning mean final acquired
  corruption.
  \item It identifies a metric reversal on clean MAGIC data: uncertainty
  sampling improves balanced accuracy while reducing average precision and
  TPR at every pre-specified FPR limit.
\end{itemize}

\subsection{Research Questions and Hypotheses}

The central research question is:

\begin{quote}
\textbf{How do random classification noise and bounded
difficulty-dependent label noise affect the relative performance of
uncertainty sampling and random sampling across annotation budgets, and do
performance differences remain after aligning acquired corruption exposure?}
\end{quote}

Four hypotheses were fixed before scientific outcomes were accessed:

\begin{itemize}
  \item \textbf{H1:} Under clean labels, uncertainty sampling has higher
  normalized balanced-accuracy AULC than random sampling.
  \item \textbf{H2:} As noise increases, the uncertainty-minus-random AULC
  advantage shrinks more under difficulty-dependent noise than under RCN.
  \item \textbf{H3:} The uncertainty-minus-random performance difference
  changes across annotation budgets. This trajectory analysis was
  pre-specified as descriptive and received no pointwise hypothesis tests.
  \item \textbf{H4:} At budget 120, uncertainty sampling under
  difficulty-dependent noise performs worse than under RCN matched to the
  fraction of corrupted labels it acquires.
\end{itemize}

\section{Related Work}

\subsection{Active Learning and Evaluation Across Budgets}

Lewis and Gale demonstrated that sequentially selecting examples could reduce
the labeled data needed for text classification \citep{lewis1994}. Settles
later organized uncertainty, disagreement, and representativeness methods
within the broader active-learning literature \citep{settles2009}. However,
an active method's measured advantage can depend on training choices,
randomness, dataset, and budget. Munjal et al.\ showed that regularization and
training settings can change active-learning conclusions \citep{munjal2022},
while Ji et al.\ emphasized repeated, controlled evaluation
\citep{ji2023}. Bae et al.\ further showed that a method effective at one
label budget can underperform random sampling in another \citep{bae2025}.
These findings motivate paired seeds, budget-dependent regularization
selection, and complete learning-curve analysis.

\subsection{Random and Input-Dependent Label Noise}

Under RCN, each binary label is independently flipped with one constant
probability below one half \citep{angluin1988}. Massart noise permits this
probability to depend on the input while retaining an upper bound below one
half \citep{massart2006,awasthi2015}. General Massart noise can be adversarial.
The present study uses a narrower, non-adversarial process in which error
probability is a fixed monotone function of difficulty relative to an
estimated reference boundary. Because that boundary is not the Bayes boundary
and need not match the active learner's boundary, the process is called
\emph{bounded difficulty-dependent noise}. It is inspired by bounded
input-dependent noise but is not claimed to simulate arbitrary Massart noise
or inherit theoretical guarantees from that setting.

Recent work has addressed active learning with imperfect annotation.
Nuggehalli et al.\ study active learning when imbalance and label noise occur
together \citep{nuggehalli2025}. Shafir et al.\ propose a noise-aware
low-budget framework and evaluate multiple noise processes on image
benchmarks \citep{shafir2025}. The present experiment complements this work
with standard logistic regression, tabular datasets, explicit exposure
matching, and paired comparisons across the complete annotation path.

\subsection{Metrics Under Class Imbalance}

AUROC averages performance across false-positive operating regions that may
not match actual use. Precision--recall analysis can expose differences that
appear small in ROC space, particularly with class imbalance
\citep{davis2006,saito2015}. This study therefore excludes AUROC and instead
reports balanced accuracy, average precision, and true-positive rate (TPR) at
pre-specified false-positive-rate (FPR) limits. MAGIC is additionally
evaluated at 1\% and 2\% FPR because its documentation states that accepting a
background event as signal is more costly than rejecting a signal event and
identifies 1\%, 2\%, 5\%, and 10\% as relevant operating thresholds
\citep{bock2004}.

\section{Methods}

\subsection{Datasets and Splits}

Three public UCI binary-classification datasets were selected to vary in size,
dimensionality, class balance, and domain (Table~\ref{tab:datasets}).

\begin{table}[ht]
\centering
\small
\begin{tabular}{lrrlcc}
\toprule
Dataset & Instances & Features & Positive class & Positive rate & FPR limits \\
\midrule
Breast Cancer \citep{wolberg1993} & 569 & 30 & Malignant & 0.374 & 5\%, 10\% \\
Banknote \citep{lohweg2012} & 1,372 & 4 & Class 1 & 0.445 & 5\%, 10\% \\
MAGIC \citep{bock2004} & 19,020 & 10 & Gamma & 0.648 & 1\%, 2\%, 5\%, 10\% \\
\bottomrule
\end{tabular}
\caption{Benchmark datasets, mean pool positive-class rates under the
stratified splits, and pre-specified fixed-FPR operating points.}
\label{tab:datasets}
\end{table}

For each of 100 seeds, each dataset received an 80/20 stratified split into an
active-learning pool and a clean test set. Numerical features were
standardized using the pool features only. Test labels were excluded from
training, query selection, cross-validation, and synthetic-noise construction.
All comparisons remained paired within dataset and seed; raw metric values
were never pooled across datasets.

\subsection{Learner and Query Strategies}

The learner was logistic regression with $L_2$ regularization, the
\texttt{liblinear} solver, and no class weighting, implemented with
scikit-learn \citep{pedregosa2011}. At every annotation budget and separately
for each query strategy, inverse regularization strength was selected from

\[
C\in\{10^{-3},10^{-2},10^{-1},1,10,100,1000\}
\]

using three-fold seeded stratified cross-validation on only the labels
acquired at that point. Cross-validation optimized balanced accuracy and
broke exact ties in favor of the smaller $C$. The model was then retrained
from scratch.

Each trial began with the same stratified set of 20 trusted clean labels,
including at least three examples from each class. Models acquired labels in
batches of five through budget 120, producing 21 recorded checkpoints.
Random sampling selected uniformly without replacement. Uncertainty sampling
selected the five remaining examples with the smallest current learner margin

\[
m_i^{\text{learner}}=\left|2\widehat p_i-1\right|,
\]

with ties broken by original pool index.

\subsection{Independent Reference Difficulty}

Synthetic difficulty-dependent error required a measure independent of
the logistic active learner. For each split, a random-forest reference model
generated one clean out-of-fold probability for every pool example using
five-fold stratified cross-fitting. Each forest used 500 trees, minimum leaf
size five, square-root feature subsampling, bootstrap sampling, and a fixed
seed rule. No example's difficulty was predicted by a forest trained on that
example. Reference difficulty was

\[
d_i=1-\left|2p_i^{\mathrm{OOF}}-1\right|,
\]

where larger values indicate proximity to the reference boundary. The
reference was frozen before any active-learning query and used only to
construct synthetic annotation error. It was unavailable to the learner and
was not used for evaluation. Consequently, $d_i$ measures proximity to the
random-forest reference boundary, not necessarily to the logistic learner's
own boundary. This distinction limits mechanistic interpretations of the
structured-noise results.

\subsection{Noise Processes}

The clean initial set was never corrupted. For the remaining pool, expected
noise rates were

\[
\rho\in\{0,.025,.05,.075,.10,.15,.20,.25,.30\}.
\]

Under RCN, every eligible label flipped independently:

\[
B_i\sim\operatorname{Bernoulli}(\rho), \qquad
\widetilde y_i=y_i\oplus B_i.
\]

Under bounded difficulty-dependent noise, eligible flip probabilities were

\[
q_i=\rho+\gamma_\rho(d_i-\bar d),
\]

\[
\gamma_\rho=0.90\min\left\{
\frac{\rho}{\bar d},
\frac{0.45-\rho}{1-\bar d}
\right\}.
\]

This centered linear construction preserves mean expected error $\rho$,
increases monotonically with difficulty, and keeps every probability below
0.45. A constant-difficulty pool falls back to RCN. Labels then flipped
independently as $B_i\sim\operatorname{Bernoulli}(q_i)$. Within each paired
condition, random and uncertainty sampling shared the same pre-generated
noise map, allowing their acquired corruption exposure to differ only because
they queried different examples. RCN and difficulty-dependent maps at the same
dataset, seed, and expected rate used deterministic but independent random
streams. H2 therefore compares independently realized noise processes rather
than applying two thresholds to one shared uniform draw.

\subsection{Exposure-Matched and Extended Controls}

H2 can reflect two mechanisms: uncertainty sampling may acquire more corrupted
labels, or corrupted labels near the boundary may be more harmful even at the
same exposure. H4 separated these possibilities with an exposure-matched RCN
control. For each dataset, source noise rate, and seed, the RCN target was the
mean final acquired-corruption fraction from the other 99 difficulty-dependent
uncertainty trials. The leave-one-seed-out target was capped at 0.45 and used
an independent noise stream. This generated 2,400 matched-control trials.
The control aligns final exposure in expectation across seeds rather than
matching it exactly within each seed, and it does not match the exposure
trajectory before budget 120. H4 therefore tests whether a residual
performance difference remains after mean final exposure alignment; it does
not perfectly isolate error location.

A secondary clean-label experiment extended Breast Cancer Wisconsin from
budget 120 to 400. Both query strategies were run for the same 100 seeds,
producing 200 trials and 35 retained checkpoints. Its first 21 checkpoints
were required to match the main experiment exactly.

\subsection{Metrics and Statistical Analysis}

The primary outcome was normalized area under the balanced-accuracy learning
curve from budgets 20 to 120, calculated by trapezoidal integration. Secondary
outcomes were final balanced accuracy, average precision, TPR at fixed FPR,
and acquired-corruption fraction. At FPR limit $\alpha$, the reported TPR was

\[
\max_{\tau:\operatorname{FPR}(\tau)\leq\alpha}
\operatorname{TPR}(\tau).
\]

The seed within dataset was the paired analysis unit. Mean paired differences
received 10,000 paired-seed percentile bootstrap resamples and 95\%
confidence intervals \citep{efron1979}. Two-sided Wilcoxon signed-rank tests
\citep{wilcoxon1945} were secondary to estimates and intervals. Holm
correction \citep{holm1979} was performed separately within H1 (3 tests), H2
(24 tests), and H4 (24 tests). H3 budget intervals and final secondary metrics
were descriptive and received no pointwise $p$-values.

H2 used the difference-in-differences

\[
D_\rho=
\left(\mathrm{AULC}_{U}-\mathrm{AULC}_{R}\right)_{\mathrm{difficulty}}
-
\left(\mathrm{AULC}_{U}-\mathrm{AULC}_{R}\right)_{\mathrm{RCN}},
\]

where negative values match the predicted direction. H4 used
difficulty-dependent minus exposure-matched RCN balanced accuracy at budget
120.
AULC robustness boundaries were defined as the first adjacent linear
zero-crossing of the mean paired uncertainty-minus-random curve over
$[0,0.30]$; absent crossings were reported as censored rather than
extrapolated.

The analysis plan and implementation were frozen internally in the repository
after all three experiment components passed count-only structural audits and
before scientific outcomes were accessed. This was a repository-recorded
freeze, not an external preregistration. The final inventory contained 12,800
trials and 271,600 retained budget rows.

\section{Results}

\subsection{H1: Clean-Label Efficiency}

\begin{figure}[ht]
\centering
\includegraphics[width=0.92\textwidth]{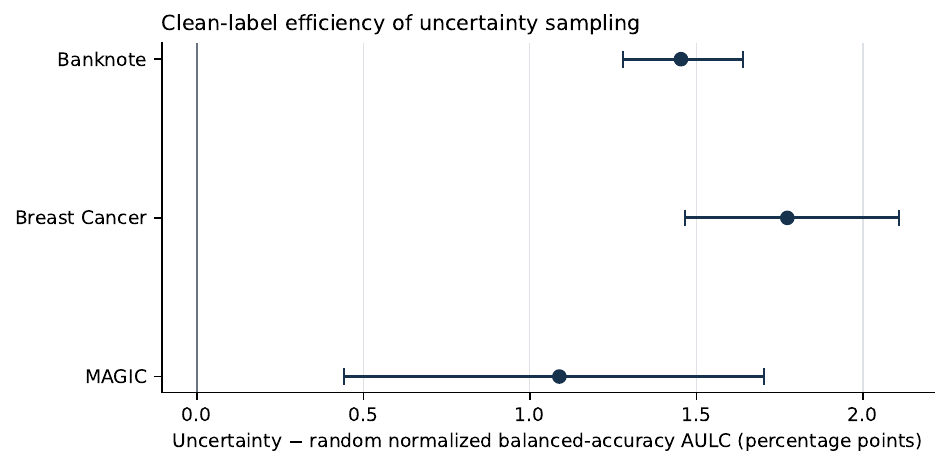}
\caption{Paired clean-label uncertainty-minus-random normalized
balanced-accuracy AULC differences. Points are means across 100 paired seeds;
bars are 95\% paired-bootstrap intervals. Positive values favor uncertainty
sampling.}
\label{fig:h1}
\end{figure}

H1 was supported on all three datasets (Figure~\ref{fig:h1}). Uncertainty
sampling increased normalized balanced-accuracy AULC by 1.45 percentage points
on Banknote (95\% CI 1.28 to 1.64; Holm-adjusted
$p=1.89\times10^{-17}$), 1.77 points on Breast Cancer (95\% CI 1.47 to
2.11; adjusted $p=5.85\times10^{-16}$), and 1.09 points on MAGIC (95\%
CI 0.44 to 1.70; adjusted $p=1.53\times10^{-4}$).

\FloatBarrier

\subsection{H2: Noise Structure}

\begin{figure}[ht]
\centering
\includegraphics[width=\textwidth]{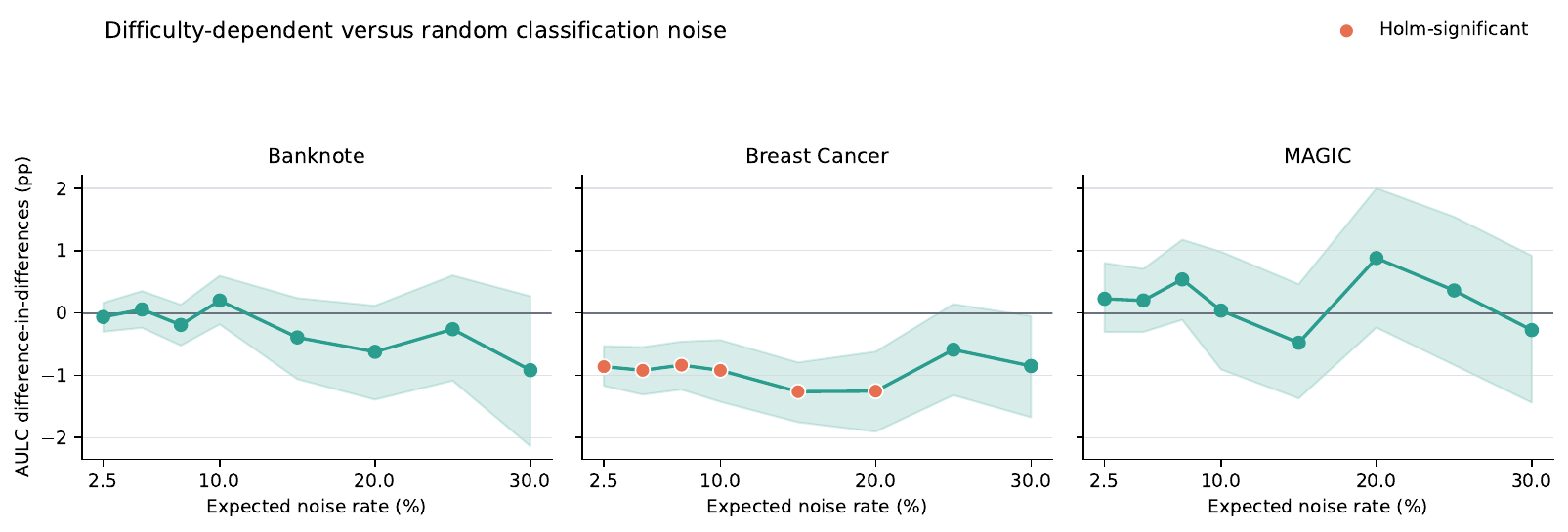}
\caption{H2 difference-in-differences in normalized balanced-accuracy AULC.
Negative values indicate that the uncertainty-minus-random advantage was
smaller under difficulty-dependent noise than under RCN. Shading shows 95\%
paired-bootstrap intervals; orange points survived Holm correction within the
24-test H2 family.}
\label{fig:h2}
\end{figure}

H2 received strong but dataset-specific support (Figure~\ref{fig:h2}). On
Breast Cancer, the difficulty-minus-RCN difference-in-differences was negative
and Holm-significant at six consecutive rates from 0.025 through 0.20. Mean
effects ranged from $-0.84$ to $-1.26$ percentage points. The 0.25 comparison
was inconclusive. At 0.30, the bootstrap interval remained below zero, but the
Wilcoxon result did not survive Holm correction.

Banknote showed no corrected difference at any rate: six of eight point
estimates were negative, but every interval crossed zero. MAGIC showed only
two negative point estimates and no corrected differences. Therefore, the
hypothesis that difficulty-dependent noise generally reduces uncertainty
sampling's advantage more than RCN was not supported across datasets.

\FloatBarrier

Despite the relative degradation on Breast Cancer, no point-estimate AULC
curve crossed below zero within the tested domain. All six dataset-process
crossing estimates were right-censored above 0.30. Thus, the results show
reduced advantage in some conditions, not general failure of uncertainty
sampling through 30\% expected noise.

\subsection{H3: Budget Dependence and the Extended Clean Curve}

Budget-level results were descriptive. Banknote's mean balanced-accuracy
advantage was positive at every post-initial checkpoint under every process
and rate. Breast Cancer remained mostly positive, but under
difficulty-dependent noise at rate 0.30 it was positive at 13 of 20
post-initial checkpoints and
ended at $-0.24$ percentage points (95\% CI $-1.33$ to 0.91). MAGIC was more
variable across budgets. No budget-level interval on any dataset was strictly
below zero.

\begin{figure}[ht]
\centering
\includegraphics[width=\textwidth]{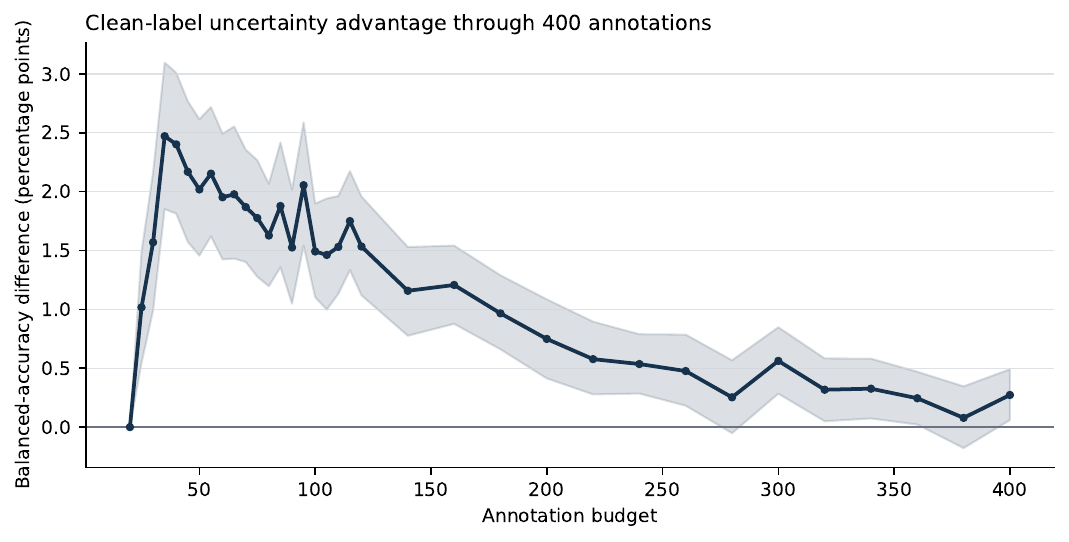}
\caption{Descriptive paired uncertainty-minus-random balanced-accuracy
difference on clean Breast Cancer Wisconsin data through budget 400. The line
is the mean across 100 paired seeds and the band is a 95\% paired-bootstrap
interval.}
\label{fig:extended}
\end{figure}

The extended Breast Cancer experiment found its first persistent positive
mean difference at budget 25, the earliest post-seed checkpoint
(Figure~\ref{fig:extended}). The mean advantage peaked at 2.47 percentage
points at budget 35, declined as random sampling caught up, and remained
positive at budget 400 (0.27 points; 95\% CI 0.06 to 0.49). The earliest
persistent-positive point estimate occurred at budget 25, but the corresponding
bootstrap interval spanned 25 to 400. The curve, rather than a precise onset
estimate, is therefore the interpretable result.

\FloatBarrier

\subsection{H4: Error Location Beyond Exposure}

The matched control aligned mean final acquired exposure closely
(Table~\ref{tab:exposure}). Across the eight nonzero source rates, every
paired-bootstrap interval for the exposure difference included zero, and the
0.45 target cap was never activated. This is evidence of successful alignment
at the dataset--rate level, not exact matching within every seed.

\begin{table}[ht]
\centering
\small
\begin{tabular}{lccc}
\toprule
Dataset & Mean exposure-difference range (pp) & All 95\% CIs include 0 & Cap uses \\
\midrule
Banknote & $-0.44$ to $+0.60$ & Yes & 0 \\
Breast Cancer & $-0.37$ to $+0.49$ & Yes & 0 \\
MAGIC & $-0.95$ to $+0.17$ & Yes & 0 \\
\bottomrule
\end{tabular}
\caption{Exposure alignment for H4 across nonzero source rates.
Differences are difficulty-dependent minus exposure-matched RCN acquired
corruption fractions at budget 120, summarized across the eight rates per
dataset. The ranges summarize means and do not establish statistical
equivalence.}
\label{tab:exposure}
\end{table}

\begin{figure}[ht]
\centering
\includegraphics[width=\textwidth]{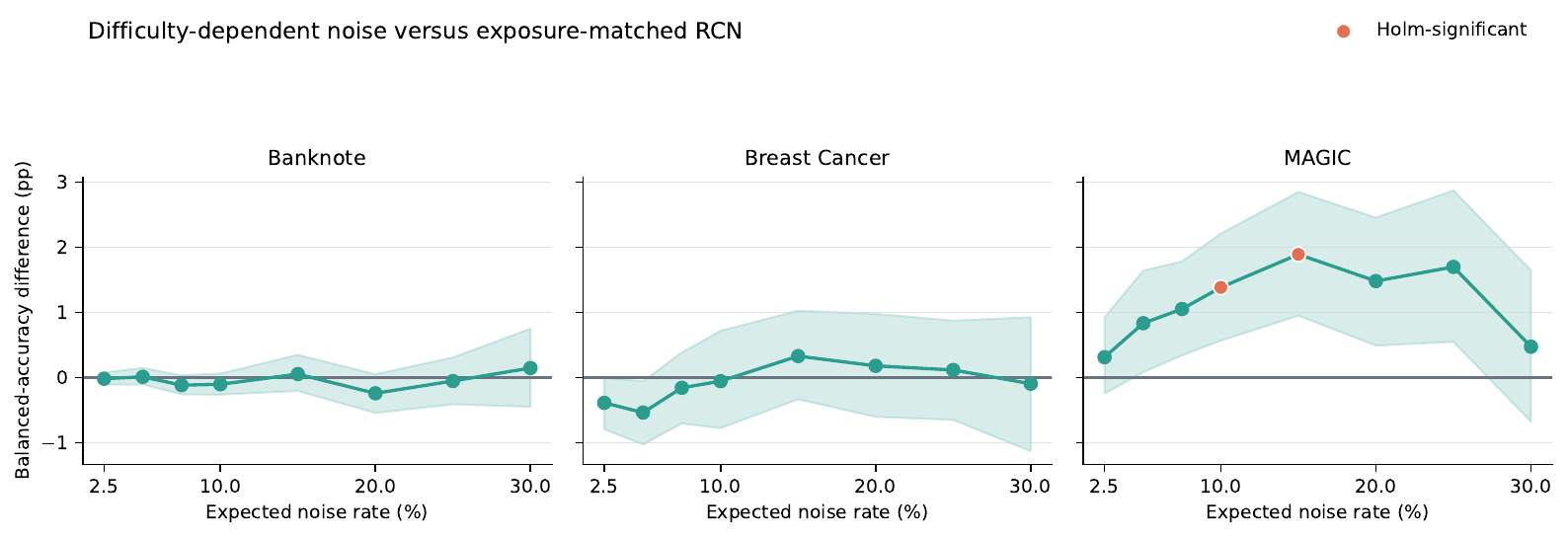}
\caption{H4 balanced-accuracy differences at budget 120. Values are
difficulty-dependent uncertainty sampling minus uncertainty sampling under
exposure-matched RCN. Negative values match H4's predicted direction.
Shading shows 95\% paired-bootstrap intervals; orange points indicate
two-sided Holm-significant results and occur in the opposite direction.}
\label{fig:h4}
\end{figure}

H4 was not supported (Figure~\ref{fig:h4}). No comparison was
Holm-significant in the predicted negative direction. Banknote remained near
zero. Breast Cancer had negative intervals at rates 0.025 and 0.05, but
neither corresponding rank test survived correction.

MAGIC instead moved mainly in the opposite direction. Difficulty-dependent
noise outperformed exposure-matched RCN by 1.38 percentage points at rate
0.10 (95\% CI 0.57 to 2.21; adjusted $p=0.0128$) and by 1.89 points at rate
0.15 (95\% CI 0.95 to 2.85; adjusted $p=0.00634$). These results reject a
simple universal claim that errors concentrated near the reference boundary
must be more harmful than randomly located errors with similar mean final
exposure. They do not establish equivalence or identify why MAGIC moved in
the opposite direction.

\FloatBarrier

\subsection{Metric Dependence on MAGIC}

\begin{figure}[!ht]
\centering
\includegraphics[width=0.74\textwidth]{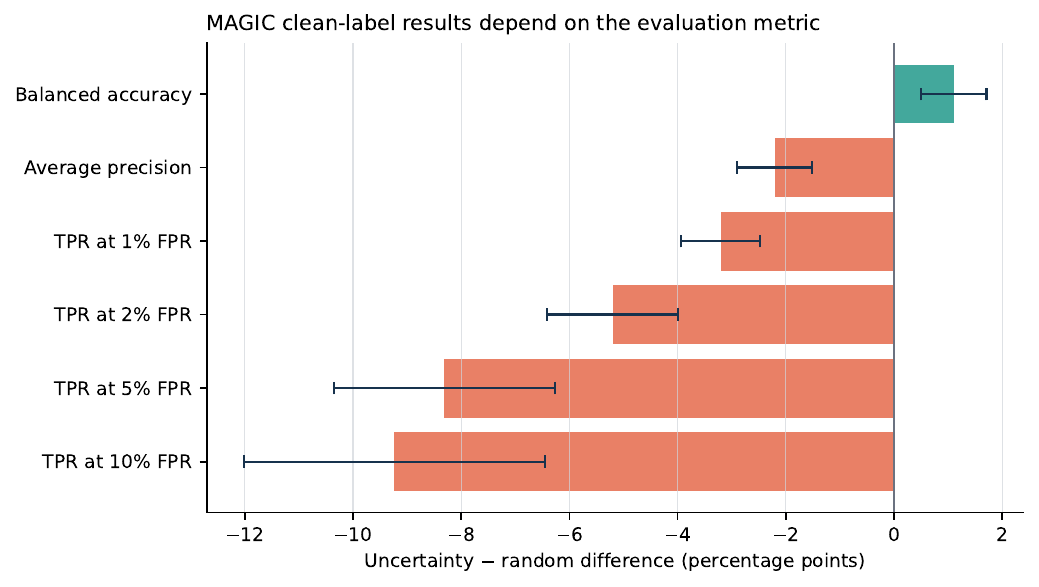}
\caption{Descriptive clean-label MAGIC differences at budget 120. Positive
values favor uncertainty sampling; negative values favor random sampling.
Bars show paired means and error bars show 95\% paired-bootstrap intervals.
These secondary intervals were not adjusted as a confirmatory test family.}
\label{fig:magic}
\end{figure}

The secondary final-budget metrics exposed a major tradeoff
(Figure~\ref{fig:magic}). Under clean MAGIC labels at budget 120, uncertainty
sampling improved balanced accuracy by 1.12 percentage points (95\% CI 0.51
to 1.71) but reduced average precision by 2.21 points (95\% CI $-2.90$ to
$-1.51$). It also reduced TPR by 3.20 points at 1\% FPR, 5.19 points at 2\%
FPR, 8.31 points at 5\% FPR, and 9.23 points at 10\% FPR. Each descriptive
interval was entirely negative.

\FloatBarrier

\section{Discussion}

\subsection{What the Results Establish}

The clean-label result is the most consistent conclusion. Across all three
datasets, uncertainty sampling used labels more efficiently according to the
pre-specified primary AULC metric. The extended experiment further suggests
that this benefit is largest when labels are scarce and narrows as the budget
grows. Random sampling is therefore a serious baseline, but it was not
stronger in the tested clean low-budget regime.

The noise comparison is less universal. On Breast Cancer,
difficulty-dependent noise reduced the active-learning advantage more than
RCN across a substantial rate range even though both processes had the same
expected pool-wide corruption rate. Banknote and MAGIC did not reproduce that
pattern. Possible explanations include dataset geometry, the alignment
between the random-forest reference and the evolving logistic learner, and
the examples reached by each query path. The current design does not
distinguish among them.

The matched control further weakens a one-mechanism explanation. After mean
final acquired exposure was aligned, the predicted additional harm did not
survive correction. MAGIC produced two corrected results in the opposite
direction. These findings do not prove that location never matters, that the
conditions are equivalent, or that exposure alone explains H2. H4 evaluated
one final budget and aligned exposure across seeds rather than within every
seed or checkpoint. The justified conclusion is narrower: a universal
additional location penalty was not detected by this control.

Finally, the MAGIC metric reversal prevents a broad statement that uncertainty
sampling was simply ``better.'' Balanced accuracy evaluates hard predictions
at a particular threshold, while average precision and fixed-FPR TPR evaluate
different aspects of ranking and operating behavior. A strategy can improve
one while reducing another. For MAGIC, the source documentation identifies
low background-acceptance regions as relevant, making the fixed-FPR results
directly important alongside balanced accuracy \citep{bock2004}.

\subsection{Practical Implications}

For small tabular projects, uncertainty sampling is a reasonable clean-label
baseline, especially when the annotation budget is very limited. It should
not be deployed with one aggregate metric and assumed to be robust. A
practical evaluation should:

\begin{itemize}
  \item compare against random sampling at every relevant label budget;
  \item tune regularization using only the labels available at that budget;
  \item repeat paired runs rather than rely on one split or query path;
  \item measure the errors actually acquired, not only the pool-wide rate; and
  \item report metrics at operating points that match the intended use.
\end{itemize}

When uncertain examples may receive unreliable labels, repeated labeling,
expert escalation, or noise-aware query rules are more defensible next steps
than assuming uncertainty sampling will either always fail or always remain
efficient.

\section{Limitations}

The conclusions are limited to three binary tabular datasets, one logistic
learner, and two basic query strategies. Deep models, multiclass problems,
structured data, and alternative uncertainty or diversity methods may behave
differently.

The noise is synthetic. The bounded difficulty-dependent process is tied to
an out-of-fold random-forest boundary, not a measurement of real human
disagreement or a worst-case bounded-noise adversary. The reference boundary
need not match the logistic learner's boundary, so ``difficulty'' is a
simulation construct rather than an observed annotator property. The
benchmark labels may themselves contain unknown errors. The trusted clean
seed of 20 labels is useful for stable cross-validation but may not exist in
every application.

Expected noise was tested only through 0.30. Since all point-estimate AULC
crossings were right-censored, this study cannot estimate where an actual
failure crossover would occur. The matched control tested its confirmatory
comparison only at budget 120 and matched mean exposure across seeds rather
than exact exposure within each seed or checkpoint. It therefore cannot fully
separate exposure from location across the trajectory. Cross-validation also
operated on very small labeled samples at early budgets; re-selecting
regularization was more realistic than fixing it, but early selections may
still be variable.

Fixed-FPR estimates depend on finite test sets and are especially discrete at
very low FPR. The MAGIC test splits contained 1,338 background examples, so a
1\% empirical FPR limit permits at most 13 false positives. Secondary metric
intervals and budget trajectories were descriptive rather than
multiplicity-adjusted confirmatory families. Finally, the analysis plan was
frozen after implementation-only pilots and structural audits rather than
registered publicly before all computation. The repository records this
sequence, but an external preregistration would provide stronger protection
in a future study.

\section{Conclusion}

Uncertainty sampling improved clean-label balanced-accuracy label efficiency
on all three tested datasets and retained a positive mean AULC advantage
through 30\% expected noise. Difficulty-dependent noise reduced this advantage
more than RCN on Breast Cancer, but the effect did not generalize to Banknote
or MAGIC. Exposure-matched controls provided no corrected evidence for the
predicted universal additional harm of structured error location and produced
two opposite-direction results on MAGIC. This null pattern does not prove
equivalence or an exposure-only mechanism. MAGIC also showed that
balanced-accuracy gains can coexist with reductions in average precision and
fixed-FPR TPR.

The defensible conclusion is therefore conditional: uncertainty sampling can
be label-efficient, but its apparent robustness depends on dataset structure,
annotation budget, noise process, and evaluation metric. Active-learning
claims should be made across complete learning curves and realistic operating
points, not from a single final score.

\section*{Reproducibility and Data Availability}

All datasets are publicly available from the UCI Machine Learning Repository.
The frozen configuration, deterministic experiment implementation, structural
audits, statistical analysis plan, automated tests, and reporting code are
available at
\url{https://github.com/dev-juy/hard-cases-bad-labels}. The retained main,
matched-control, extended-clean, and analysis artifacts are hash-bound to
manifests. The scientific analysis contains 12,800 trials and 271,600
retained budget rows. The main and matched-control components used 100 seeds
for each dataset; the extended clean-label component used 100 paired seeds on
Breast Cancer Wisconsin only.

\section*{Deviations and Reporting Clarifications}

The frozen configuration retained the internal identifier
\texttt{margin\_dependent\_massart\_style}. The manuscript uses
``bounded difficulty-dependent noise'' because the simulated probability
depends on distance to a random-forest reference boundary rather than the
logistic learner's margin, and because the process is narrower than general
Massart noise. This is a terminology clarification; the noise function,
trials, hypotheses, and statistical tests were not changed.

The H1--H4 numbering is preserved from the frozen analysis plan. The exposure
range in Table~\ref{tab:exposure} is a post-analysis descriptive summary of
the already-produced H4 output; it introduces no additional test. No new
scientific trials or post-outcome model variants were added during manuscript
revision.

\section*{Acknowledgments}

The author thanks Daniel Kane for feedback connecting the simulated processes
to random classification and Massart-style noise, Christian Shelton for
feedback on budget-dependent regularization and operating-point metrics, and
the Summit Research Scholars teaching fellows and peer reviewers for comments
on runtime, reference-model independence, exposure matching, and budget
coverage. These contributors did not determine the final analyses or claims,
and any remaining errors are the author's.

{\small
\setlength{\bibsep}{0.25em}
\bibliographystyle{unsrtnat}
\bibliography{references}
}

\end{document}